**A Monolithic Hand with Asymmetric Origami Bending and Dual-chamber Actuators**

Nan Huang [a], Yuming Zhu [a], Zicong Zhang [a], Jianhui Liu [a], Xiaohuang Liu [a], Dihan Liu [c], Jiansheng Dai [a,*], Sicong Liu [b,*]

[a] *Department of Mechanical and Energy Engineering, Southern University of Science and Technology, Shenzhen, Guangdong, 518000, China.*

[b] *Sino-German College of Intelligent Manufacturing, Shenzhen Technology University, Pingshan, China.*

[c] *University of Pennsylvania, Philadelphia, PA 19104, USA.*

## Abstract

The passive adaptability inherent in soft robotic hands affords them advantages in applications that require safe and compliant interaction. However, existing soft robotic hands often struggle to simultaneously achieve adequate output performance and easy manufacturing due to their complicated structures. In this paper, we introduce the asymmetric origami bending (AOB) pattern for generating bending motion and the asymmetric dual-chamber (ADC) design for obtaining multifunction capability. The AOB single (AOB-S) chamber and AOB dual-chamber (AOB-D) units are designed and constitute the finger and palm actuators of the proposed Origami-inspired SOft Robotic (OSOR) hand. The OSOR hand achieves bio-inspired fingers-palm motions and adequate output performance within a monolithic structure that significantly simplifies the manufacturing process. By defining the asymmetric ratio to characterize the geometric asymmetry of the unit, the analytical models of the AOB and ADC structures are proposed. The Finite Element Analysis tool for the design of AOB actuators is obtained by geometric analysis. The asymmetric origami design grants the integrated manufacturing of the OSOR hand through a Selective Laser Sintering printing process with a single thermoplastic polyurethane material. The model and simulations are validated by experimental results. Experiments show the finger and palm maximum bending motion range of 203° and 40°, respectively, with output forces of 6.3 N and 16 N. The OSOR hand is capable of pinching a piece of tissue, stably grasping water bottles with two fingers, palm-only grasping, and completing the power grasps in the taxonomy of manufacturing grasps. The compactness, performance, and easy manufacturing of the proposed hand benefit the development of the soft robotic hand with new possibilities.

# I. Introduction

Leveraging the inherent compliance and adaptability, soft robotic hands provide a safe and robust solution besides conventional robotic hands for grasping objects of irregular shapes and varying fragility, making them particularly suitable for direct human-robot interaction [1]–[21]. Compared with the rigid counterparts, soft robotic hands are lighter, simpler in structure, and excel at adapting to irregular shapes when grasping. As the motion-generating components, soft bending actuators predominantly determine hand performance and have received extensive exploration. These pioneering researches have advanced the output performance and applications of soft robotic hands [22]–[34].

Various efforts were devoted to designing soft bending actuators with adequate range of motions and load-bearing capacities for soft robotic hands, including the following representative solutions: (1) the intrinsic air chamber actuator (IACA) [35]–[38], which achieves large bending angles by pressurizing specially shaped air chambers at the joints, made from super-elastic materials such as silicone; (2) the strain-limiting actuator (SLA) [39]–[45], which achieves high load-bearing capacity by constraining deformations and guiding movements to improve energy efficiency; (3) the soft-rigid coupled actuator (SRCA) [46]–[48], which achieves high load-bearing capacity by increasing the stiffness of the actuator and guiding the output force in the desired direction. Advancements in soft actuators benefit the development of soft robotic hands in output performance. The biomimetic hand integrates 3 IACAs per finger, achieves a bending angle of about 75° per joint, and enables compliant grasping [36]. The RBO Hand 3 utilizes SLAs to deliver high fingertip output force while maintaining an extensive kinematic workspace [42]. The multi-joint hand mitigates the negative impact of soft material on the load-bearing capacity of the fingers by integrating rigid and soft materials, thereby enhancing the output force [46].

Although these advanced actuators improved output performance, robotic hands' fabrication remains challenging due to the delicate structure. Through the cooperation of actuator design and additive fabrication methods, feasible approaches for simplified fabrication have been put forward. By adopting IACAs and multi-material 3D printing, monolithic fabrication of each finger and palm separately with distributed pneumatic tactile sensors was achieved, simplifying the assembly process of the hand [35]. A printable soft robotic hand was produced using pouch motors and planar fabrication technology [37]. The adoption of IACAs and selective laser sintering (SLS) technology enabled the monolithic fabrication of the soft robotic hand with embedded pneumatic actuators [49].

Despite progress in simplifying the fabrication process using 3D printing, a hurdle persists: achieving one-step fabrication of soft robotic hands with flexible fingers and supportive palm structures that provide adequate stiffness. The compliance of super-elastic materials endows fingers with adaptability to objects of various shapes; however, the compliance conflicts with the higher stiffness required for the palm, which supports the structures and the functions of fingers. Researchers fabricate dexterous fingers from super-elastic materials while employing stiffer materials for the palm [35]–[44], [46]–[48], which has yet to realize manufacturing the hands with a single material in a one-step process. Although a pioneering research has successfully achieved the fabrication of the monolithic soft robotic hand [49], the degrees of freedom (DOFs) and dexterity are limited due to the structural design of the soft actuators.

The origami-inspired soft pneumatic actuator (OSPA), which achieves desired outputs by the design of folding creases, is considered a promising approach to realizing specific motions and excellent output performance within a monolithic structure [50]–[70]. By arranging the creases into a desired pattern, OSPA achieves motion and displacement programming [56]–[81]. The Yoshimura pattern imparts linearity in axial extension and contraction, thereby improving motion stability, with typical applications encompassing wearable prosthetic and robotic arms [53], [82], [83]. The Pleated pattern achieves axial contraction under positive pressure, effectively augmenting load-bearing capacity during contraction, with applications including grippers and soft hands [84], [85]. The Kresling pattern introduces a unique spiral folding mode that enables torsional motion in OSPAs, making it suitable for applications including robotic arms and pipeline robots [86], [87]. While OSPAs are developed for diverse application scenarios, research on the OSPA designed for joint’s bending motion in soft robotic hands remains limited.

To achieve multiple-DOFs and multi-functionality with a compact structure, multi-chamber designs in OSPAs are investigated, focusing on the spatial arrangement of the chambers and the structural design of the inner walls [50]–[52], [88]–[93]. A nested dual-chamber actuator with coaxially arranged Yoshimura and Pleated chambers achieves bidirectional payload enhancement via compounding pressure actuation [50]. A dual-chamber actuator achieves large-range motions through a straight-curved crease prismatic design and a foldable internal partition [51]. A multi-chamber actuator with foldable inner walls achieves the integration of three-directional motions and proprioception [52]. These approaches demonstrate the potential of the multi-chamber design in OSPAs. Nevertheless, the multi-chamber design for the bending OSPAs still requires investigation.

In this study, we develop an asymmetric origami bending (AOB) pattern capable of generating the bending motions for finger and palm joints. The introduction of the asymmetric dual chamber (ADC) design within the pneumatic chamber enables the finger actuator with 2-DOF bending motions. Combining AOB pattern and ADC design in the actuators, an Origami SOft Robotic (OSOR) hand is produced through monolithic 3D-printing. The main contributions are summarized as follows:

1. The asymmetric origami bending (AOB) pattern is proposed based on the Yoshimura pattern for bending motions in the OSOR hand. The asymmetric dual-chamber (ADC) design was introduced in the AOB pattern for obtaining multifunctionality within the chamber. The dual-chamber (AOB-D) and single-chamber (AOB-S) origami units are obtained and constitute finger and palm actuators of the hand, generating 12-DOF motions.
2. Analytical models on the quasi-static behavior of the AOB pattern and ADC design were established, which are validated through experiments. An asymmetric ratio is proposed to characterize the geometric asymmetry of the chamber unit, upon which the motions of AOB chambers are analyzed. To obtain desired performance, comprehensive Finite element analyses (FEAs) are conducted to investigate the geometric parameters in the actuators, obtaining a simulation tool for the design.
3. Enabled by the design, the OSOR hand with 5 fingers and a palm was fabricated as a single-material monolithic structure using SLS printing. The static performance, flexibility, and grasping performance of the OSOR hand were tested and compared with state-of-the-art works, proving the OSOR hand’s advantage.

The structure of the article is as follows: Section II presents the concept of AOB pattern, ADC design and the OSOR hand; Section III details the AOB actuators' structure, establishes analytical model, and the programmable performance of AOB and ADC was investigated; Section IV reports the fabrication process and detailed design of the OSOR hand; The validation of the model and performance of the OSOR hand is introduced in Section V; the Conclusion follows.

## II. Concept of AOB pattern and OSOR hand

In this section, we introduce the concept of AOB pattern and the OSOR hand, and explain how to derive the corresponding actuators from AOB pattern to form the OSOR hand.

### A. Asymmetric origami bending pattern

Bending or twisting motions can be obtained by purposefully eliminating specific symmetries of an origami structure, from which the AOB pattern is proposed as shown in Fig. 1(a). The Yoshimura origami pattern with triple planar symmetry generates displacement governed by three symmetry planes parallel to plane XOY, YOZ, and ZOX, respectively, thereby limited to linear motions. By selectively eliminating a specific symmetry plane, the AOB pattern obtains asymmetric displacement from the folding creases on either side of the eliminated plane during deployment, thereby transforming the linear motion into other motions.

As shown in Fig. 1(a), the planar symmetric origami prismatic tubes can be defined as a lofted geometry that takes the plane normal to the direction of motion as its symmetry plane, and their spatial structure can be regarded as being formed by lofting from the end plane to the middle plane. The end plane refers to the two planes at the extremities of the origami tube, and the middle plane typically denotes the plane equidistant from both end planes, both of which are typically formed by alternating mountain and valley creases. By changing the relative pose of the end and middle planes, or increasing the number of creases, the symmetry of the origami tube is changed.

Three geometric altering operations were employed to derive functionally distinct foldable units: (1) displacing the center point of the end plane to cancel the symmetry about the YOZ or XOY planes, thereby generates unit i and unit ii capable of bending around Z-axis or X-axis, respectively; (2) twisting the end plane to remove the symmetry about the YOZ and XOY planes simultaneously, thus generates unit iii capable of a twisting motion; (3) increasing the number of foldable mountains within single unit that eliminate symmetry about YOZ, results in bending unit iv. To mimic hand joint flexions, the AOB foldable unit i was chosen as the origami actuation structure for the OSOR hand, given its compact profile and motion type.

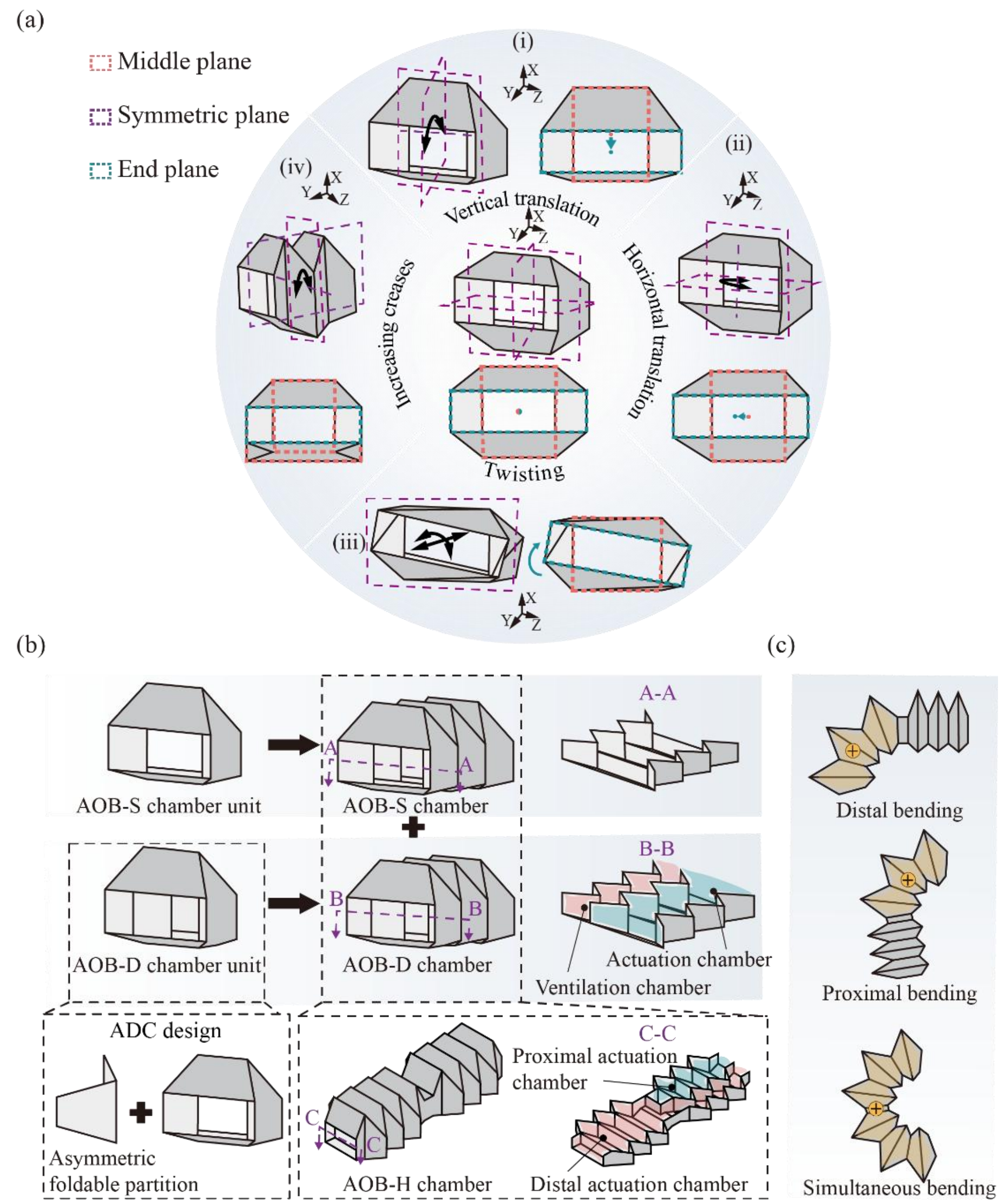

Fig. 1 The concept of the AOB pattern and ADC design. (a) The concept of the AOB pattern involves altering the symmetry of the chamber to deduce different motion types. By altering the relative positions of the end plane and the middle plane or increasing mountain creases to break the symmetry in different directions of the chamber, we obtained chambers with bending or twisting motion. (b) The ADC design involves introducing an asymmetric foldable partition to the chamber, which results in an asymmetric dual-chamber structure. Derived from the AOB pattern and ADC design, the AOB-S chamber and AOB-D chamber are obtained. By connecting them in series, the AOB-H chamber with the proximal actuation chamber and distal actuation chamber is formed. (c) Three motion modes of the AOB-H chamber are obtained by pressurizing different chambers

B. Asymmetric dual-chamber design

By introducing an asymmetric foldable partition into the AOB foldable unit, the asymmetric dual-chamber design (ADC) is proposed, which forms a dual-chamber AOB structure for multi-function usage. As shown in Fig. 1(b), the AOB-D chamber comprises two segregated and functionally specialized chambers: (1) a ventilation chamber responsible for air exchange, and (2) an actuation chamber which actively generates deformation via pressurization. The asymmetry of the partition grants each chamber different behaviors subjected to pressure changes. By serially connecting the AOB-D and AOB-S chambers, specifically by linking the ventilation chamber of the AOB-D to the AOB-S to form an integrated chamber while the actuation chamber of the AOB-D remains independent, we introduce the AOB hybrid (AOB-H) chamber, which includes the proximal and distal actuation chambers. The AOB-H chamber achieves three motion modes: proximal bending is obtained when the proximal actuation chamber is pressurized, distal bending occurs when the distal actuation chamber is pressurized, and simultaneous bending is achieved when both chambers are pressurized simultaneously, as shown in Fig. 1(b) and (c).

C. Finger and palm actuator

By implementing AOB-H and AOB-S designs, respectively, we have developed finger actuators and palm actuators for the OSOR hand, as shown in Fig. 2 (a) and (b). Each finger actuator is designed with 2-DOF using the AOB-H design, which is composed of an AOB-S chamber at the distal joint and an AOB-D chamber at the proximal joint, as shown in Fig. 2(a). The ventilation chamber of the AOB-D is adopted as the pneumatic channel of the distal joint. The actuators for the palm joints are both constructed by the AOB-S chamber with increased width for obtaining greater bending torque output, as shown in Fig. 2(b). In view of the fabrication process (described in Section IV), the finger actuators keep the external crease structure but use an internal non-crease structure on the short mountain-crease side, as illustrated in Fig. 2(a). This design is also applied to the palm actuators.

D. Bioinspired OSOR Hand

To achieve human-mimic dexterity in the soft robotic hand, the anatomical structure of human hand is used as the inspiration, as shown in Fig. 2(c). We designate the human hand structure into four distinctive functional kinematic regions for designing the corresponding structures in the OSOR hand: the fingers region encompassing the phalangeal segments [45], the left palm region formed by the metacarpals V and IV, the middle palm region constituted by metacarpals III and II, and the right palm region primarily consisting of the metacarpals I as shown in Fig. 2(c-i).

For the finger design, we analyzed the human finger anatomy and employed bending motions of AOB-H to mimic the joint flexions induced by the flexor pollicis longus and brevis, achieving two bending DOFs in each finger. As shown in Fig. 2(c-ii), when positive pressure is supplied to the distal actuation chamber, the finger actuator produces the distal bending, emulating the flexion effect of the flexor pollicis longus. When positive pressure is supplied to the proximal actuation chamber, the actuator produces proximal bending, emulating the flexion effect of the flexor pollicis brevis.

For the palm design, AOB-S actuators are used to mimic the joint flexion induced by the opponens digiti minimi and opponens pollicis muscles. When positive pressure is applied to the palm actuator in the left palm, the AOB-S chamber undergoes bending, driving the left palm to form a folding motion which simulates the flexion effect of the opponens digiti minimi. When positive pressure is applied to the palm actuator in the right palm, the AOB-S chamber produces the opposition motion simulating the flexion effect of the opponens pollicis, as shown in Fig. 2(c-iii).

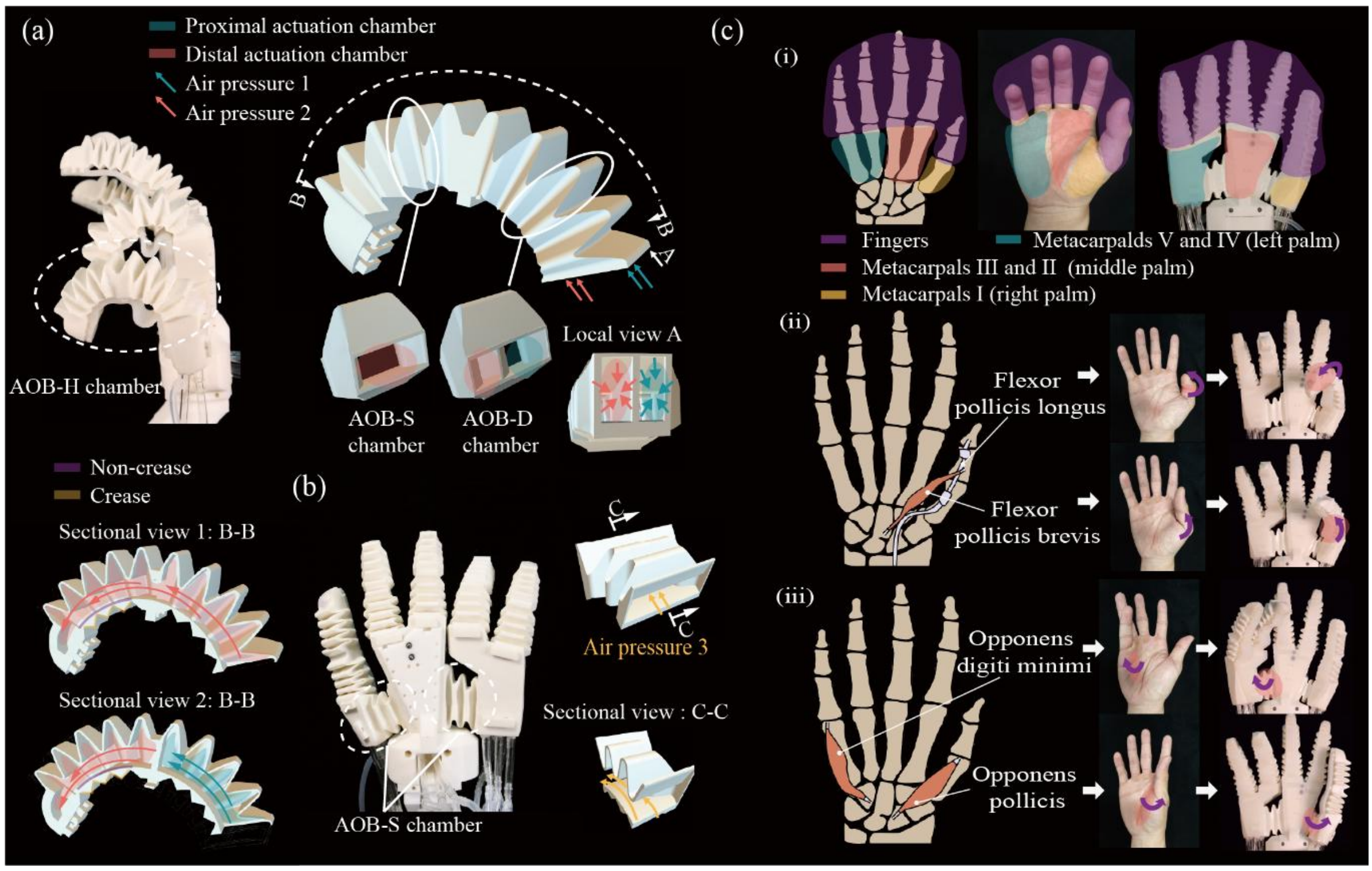

Fig. 2 (a) Finger actuators and (b) palm actuators of the OSOR hand are constructed based on the AOB-H chamber and AOB-S chamber, respectively. The independent distal and proximal actuation chambers of the finger actuator generate 2-DOF motions. (c) Bio-inspired soft hand concept divides the hand into fingers, left palm, middle palm, and right palm region. Using finger actuators and palm actuators, the motion modes resulting from thumb muscle contraction and palm muscle contraction are simulated.

## III. Modeling and Simulation

The design of AOB actuator plays a vital role in the performance of the OSOR hand. The analytical models of the actuators are used to predict mechanical behaviors and provide design guidance for geometric parameters. In this section, we derive the quasi-static model of the AOB actuator and give a simulation tool for the AOB actuators design.

### A. Analytical Model of AOB pattern

The AOB actuator is formed by connecting AOB foldable units in series, and sealing the ends to create the pneumatic chamber. Under positive pressure loading, a typical AOB actuator increases chamber volume through unfolding of mountain and valley creases. Inspired by small strain folding (SSF) principle [53], the folded portion of the facet is treated as a cantilever beam fixed at the crease, with the bending localized near the crease, where the width of the bending region is proportional to the thickness of the origami facet. Using the principle of virtual work, a relationship can be established when the actuator is subjected to a moment

$$M_i \cdot \frac{\mathrm{d}\beta'}{m} = \sum M_f \cdot \mathrm{d}\theta_f + \sum f_s \cdot \mathrm{d}l_s, \tag{1}$$

where $M_i$ represents the externally applied moment, $\beta'$ represents the corresponding rotation angle of the actuator, and $m$ denotes the number of units connected in series in the actuator. $M_f$ and $\theta_f$ represent the out-of-plane bending moment and rotation angle of creases, respectively, while $f_s$ and $l_s$ represent the in-plane axial force and the corresponding axial elongation, respectively. To better describe the relationship between the geometric parameters and the deformation characteristics of the AOB chamber, a kinematic model needs to be established.

For the actuator with $m$ AOB units, the foldable unit exhibits high programmability and can be regarded as a lofted geometry from the end plane to the middle plane, as shown in Fig. 3(a). Therefore, it includes the following design parameters: $a_1$, $b$, $a_2$, $b_1$, $h$ and $t$, the definitions of which are listed in Table S1 of the Supplementary Information 1 (SI 1). To describe the asymmetry in the AOB pattern for the generation of bending motions, the asymmetric ratio $R$ is defined as the ratio of the distance $v$ (between the centers of the rectangular end plane and middle plane) to the disparity $b_1$ - $a_1$, symbolizing the offset distance of the end plane relative to the middle plane. The range of $R$ is [0, 0.5]. When $R = 0$, the structure reverts to the Yoshimura pattern, and as $R$ increases, the degree of asymmetry rises. Then, these 7 basic parameters ($a_1, a_2, b_1, b_2, h, t, R$) define an AOB unit, as shown in Fig. 3(a).

During the deformation, the planes $B_1B_2C_1C_2$ and $E_1E_2F_1F_2$ rotate around the valley creases $B_1B_2$ and $C_1C_2$, respectively. Assuming the lengths of the creases $B_1C_1$, $B_2C_2$, $E_1F_1$ and $E_2F_2$ remain unchanged, and to ensure deformation compatibility, the planes $C_1C_2D_1D_2$ and $D_1D_2E_1E_2$ undergo corresponding tensile deformation, as shown in

Fig. 3(b). The deformation of planes $A_1A_2B_1B_2$ and $A_1A_2F_1F_2$ is negligible, and the angle $\gamma(x)$ is a function that varies with $x$, which can both be validated by the FEA, as shown in

Fig. 3(c). The detailed settings of FEA are introduced in SI 2. The static model of the AOB-S is derived (see SI 3) as follows:

$$M_i \cdot \frac{\beta'}{m} = \frac{Et^2(a_2(\theta_1' - \theta_1^0)^2 + 4c_1(\varphi_1' - \varphi_1^0)^2 + b_2((\frac{\beta'}{m})^2 + (\beta' + \theta_1' - \theta_1^0)^2)}{48n} + \int_{x_0}^{x_1} \frac{Et^2(\gamma(x) - \gamma_{\min})^2}{24n}\,\mathrm{d}x$$
$$Et(a_2 + b_2)(l' - c_1 \sin\alpha_3 - c_1 \sin\alpha_3 \ln\frac{l'}{c_1 \sin\alpha_3}) + + \int_0^{b_1} \frac{Et^2(\gamma(x) - \gamma_{\min})^2}{24n}\,\mathrm{d}x, \tag{2}$$

where $n$ is a coefficient characterizing the width of the bending region obtained by experiment [53].

The external moment $M_i$ of the AOB structure under quasi-static conditions can be derived from the virtual work between the work done by the driving moment and the energy produced by the pressure, which can be expressed as

$$M_i = p\frac{\mathrm{d}V_{AOBS}}{\mathrm{d}\beta'}, \tag{3}$$

where $p$ denotes the input pressure for the AOB-S actuator and $V_{AOB\text{-}S}$ denotes the volume of the AOB-S actuator. The expression of the AOB-S actuator volume can be derived as

$$V_{AOB-S} = \frac{1}{2}ml' \cos\frac{\theta_1'}{2}(h + 2a_1 \cdot \sin\frac{\beta'}{2m})(\frac{2b_2}{3} + \frac{a_2}{3} - \frac{2}{3}\sin\alpha_4 \cdot \cos\frac{\gamma_{\max}'}{2} \cdot c_2) + ma_1b_2h -$$
$$m\int_{x_0}^{x_1} \frac{1}{2}\sin\gamma(x) \cdot (c2 \cdot \sin\alpha_4)^2\,\mathrm{d}x. \tag{4}$$

B. Analytical Model of ADC design

The introduction of the partition structure in AOB-D increases the stiffness of the chamber. As shown in

$$M_i \cdot \frac{\beta'}{m} = \frac{Et^2(a_2(\theta_1' - \theta_1^0)^2 + 6c_1(\varphi_1' - \varphi_1^0)^2 + b_2((\frac{\beta'}{m})^2 + (\beta' + \theta_1' - \theta_1^0)^2)}{48n} + \int_{x_0}^{x_1} \frac{Et^2(\gamma(x) - \gamma_{\min})^2}{16n} dx$$
$$Et(a_2 + b_2)(l' - c_1 \sin\alpha_3 - c_1 \sin\alpha_3 \ln\frac{l'}{c_1 \sin\alpha_3}) + \int_0^{b_1} \frac{Et^2(\gamma(x) - \gamma_{\min})^2}{16n} dx. \tag{5}$$

Fig. 3(d), in the case where the partition design and orientation are identical to the ventilation chamber wall, the symmetric characteristics allows the kinematic assumptions of the AOB-S to be applied to the AOB-D modeling. This is verified through simulation: FEA of the AOB-D clarifies that when the actuation and the ventilation chamber were both pressurized, the strain distribution on the partition is highly similar to that of the ventilation chamber wall as shown in Fig. 3(e). Thus, from Eqns. (S2) - (S8), the static model of the AOB-D can be derived as follows,

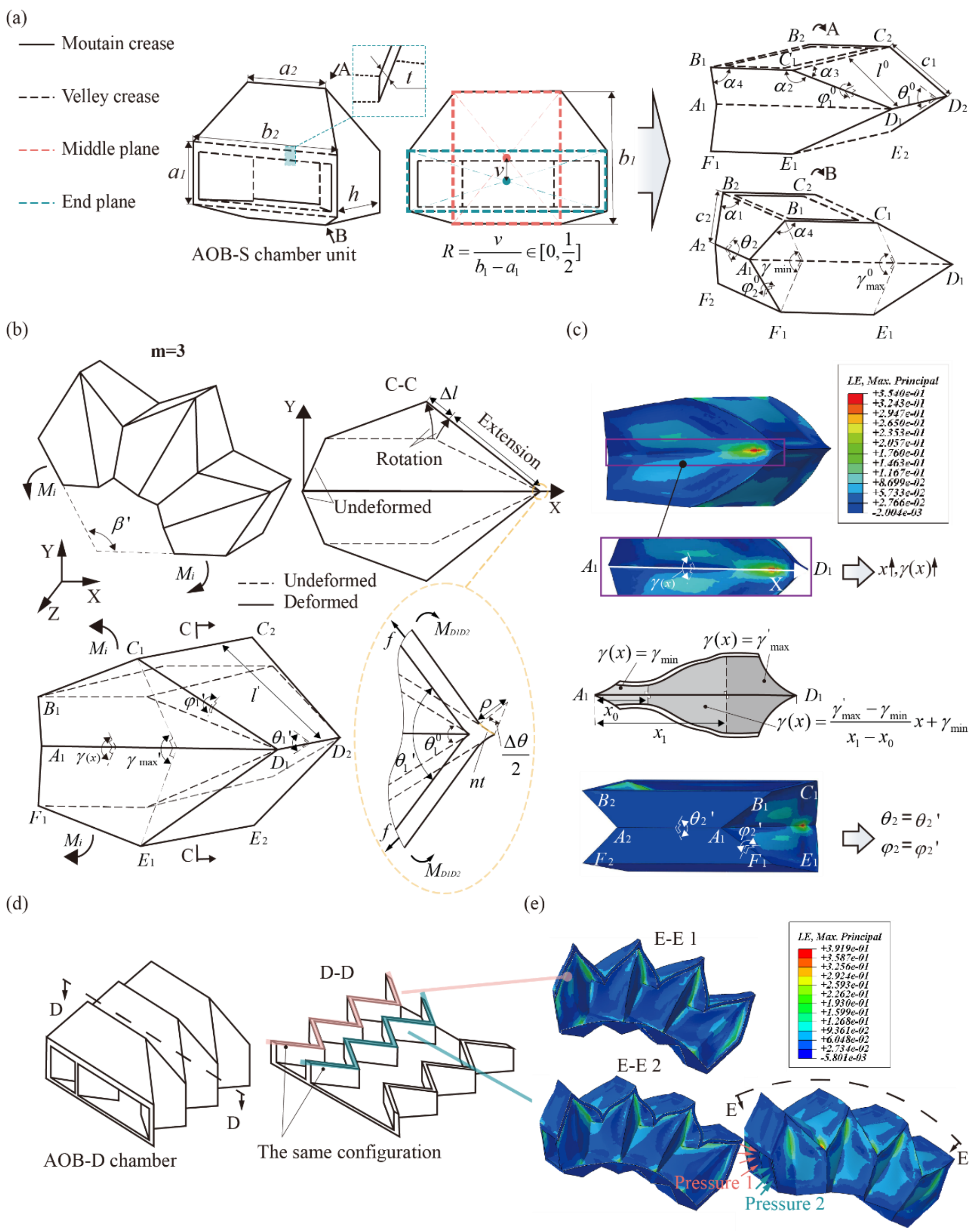

Fig. 3 Analytical model of AOB actuators. (a) Definition of 7 basic parameters and the derived geometric parameters used of the AOB-S unit used for model derivation. (b) Deformation process and assumptions for the AOB-S chamber. (c) FEA results of the AOB-S actuator verify the validity of the basic deformation assumptions. Partition configuration (d) and strain distribution (e) of the AOB-D chamber, where the partition exhibits a strain distribution analogous to that of the outer wall of the ventilation chamber.

And the volume of AOB-D actuator can be derived as

$$V_{AOB-D} = \frac{1}{2} m l' \cos\frac{\theta_1'}{2}(h + 2a_1 \cdot \sin\frac{\beta'}{2m})(\frac{2b_2}{3} + \frac{a_2}{3} - \frac{2}{3}\sin\alpha_4 \cdot \cos\frac{\gamma'_{\max}}{2} \cdot c_2) + m a_1 b_2 h -$$
$$m\int_{x_0}^{x_1} \frac{1}{2} \sin\gamma(x) \cdot (c2 \cdot \sin\alpha_4)^2 dx - mt \cdot [a_1 + (b_1 - \frac{h}{2\tan\frac{\theta_2^0}{2}})] \cdot \sqrt{\left(\frac{h}{2}\right)^2 + \left(\frac{b_2 - a_2}{2}\right)^2}. \tag{6}$$

AOB-H actuator is formed by AOB-S and AOB-D actuators in series, the static model can be obtained from Eqns. (2) - (6).

C. Parameter sensitivity analysis of AOB-S actuator

The variations of the output bending angle of the actuator were simulated under the influence of the seven basic parameters while the input air pressure increased linearly from 0 to 100 kPa. The sensitivity analysis of the basic parameters reveals distinct response patterns in the bending angle $\beta'$. Within the investigated ranges, $a_1$ and $b_1$ exhibit monotonic decreasing and monotonic increasing influences on $\beta'$, respectively. When $b_1$ is set to 25 mm, the actuator demonstrates rapid low-pressure response characteristics, as illustrated in Fig. 4(a) and (c). $a_2$, $h$ and $t$ share similar effects on $\beta'$, showing an initial increase followed by a decrease, as shown in Fig. 4 (b), (e), and (g).

Simulation results further show that low values of $h$ and $t$ generally correspond to rapid low-pressure response characteristics. $b_2$ and $R$ exhibit a trend in which $\beta'$ increases initially with the parameter values and then levels off, as illustrated in Fig. 4 (d) and (f). Parameter $m$ shows a linear multiplication effect on $\beta'$ as shown in Fig. 4 (h), which is consistent with the derivation of Eqn. (10). Fig. 4 (i) further illustrates the variations in the maximum logarithmic strain (Max LE) of the actuator under 100 kPa. Overall, parameters $a_1$, $a_2$ and $b_1$ exhibit relatively minor effects on the maximum strain, whereas the values of $h$, $R$, and $t$ exert more significant influences. The analyses provide a basis for determining the final geometric parameters of the actuator.

To quantify the influence of sensitivity of each geometric parameter on the bending angle,

a geometric sensitivity coefficient (*GSC*) is introduced, defined as

$$GSC = \frac{\beta'_{max} - \beta'_{min}}{p_{vr}}, \tag{7}$$

where $p_{vr}$ means the variation range of the parameter. The *GSC* values for all seven parameters are calculated accordingly as shown in Fig. 4 (a)~(g). The results shown that $a_1$, $t$ and $R$ yield high *GSC* values, indicating these parameters exert a stronger influence on the bending angle. Consequently, in selecting the geometric parameters of the actuator, priority is given to the values that yield the maximum bending angle, i.e. $a_1 = 7$ *mm*, $t = 0.8$ *mm* and $R = 0.32$. For the remaining parameters with lower *GSC* values, the values are chosen to achieve a balance between maintaining reasonably large bending angle and minimizing the LE. This result in the determination of $a_2$, $b_1$, $b_2$ and $h$, as listed in Table S3.

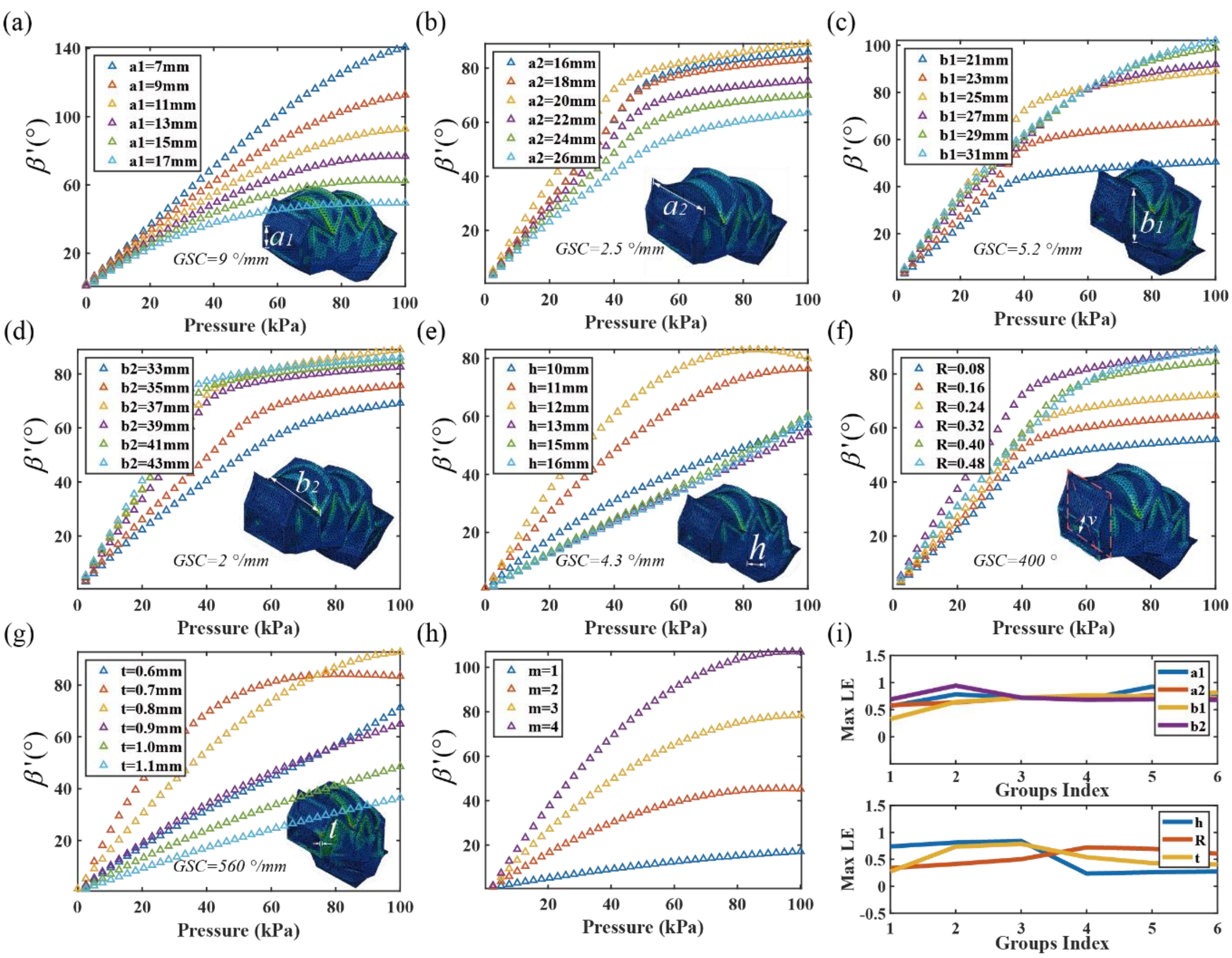

Fig. 4 Motion performance of AOB-S actuator variations. (a)~(g) Bending angle $\beta'$ of the AOB-S actuator under different variations of the 7 basic parameters. (h) The number of units in series *m* exhibits a linear relationship with the bending angle $\beta'$. (i) Summarizing the Max LE of the actuator across the six different value groups for 7 basic parameters. The horizontal axis denotes the 'Groups Index', which maps the specific values for each basic parameters, corresponding to the data presented in (a)~(g). For example, for $a_1$, groups index 1 to 6 correspond to values of 7, 9, 11, 13, 15 and 17mm, respectively.

D. Parameter sensitivity analysis of ADC design

The AOB-H actuator is formed by connecting the AOB-S and AOB-D structures in series, as shown in Fig. 5(a). To meet the requirements for 2-DOF motions, the AOB-D structure is used to form two chambers, one for actuation and one for pneumatic passage. The ADC design between the two chambers can be configured as follows: (1) identical to the proximal actuation chamber wall (ADC-A), (2) identical to the ventilation chamber wall (ADC-V), (3) a combination of both (ADC-B). FEA for the AOB-H with ADC-A revealed that when only the distal actuation chamber is pressurized to induce distal bending, an undesirable coupled motion occurs: lateral bending around the X-axis, as shown in Fig. 5(b). This unwanted deformation is generated by the constraint of the proximal actuation chamber, while the distal actuation chamber is intended to bend around the Z-axis. To mitigate this coupled displacement, two modified ADC designs, ADC-B and ADC-V, were investigated. The ADC-B design provides a symmetric constrained boundary condition. FEA shows that, due to the stiffness mismatch between the outer wall of the ventilation chamber and the partition, the reduction in coupled deformation remains limited. In contrast, the partition in ADC-V conforms to the deformation of the pressurized chamber. This design limits the coupled deformation, as shown in Fig. 5(b).

The coupling effect in the AOB-H actuator also appears in the interaction between independent chambers. Ideally, when only the proximal actuation chamber is pressurized, the end of distal actuation chamber (plane A) is expected to achieve the same bending angle as the end of proximal actuation chamber (plane B); when only the distal actuation chamber is pressurized, plane A is expected to exhibit negligible bending. To evaluate this chamber coupling effect, we simulated the $\beta^{'}$ of proximal and distal actuation chamber under three motion modes, proximal bending, distal bending and simultaneous bending, for 3 ADC designs, as shown in Fig. (d-i), (e-i), and (f-i). A chamber coupling ratio (*CCR*) was defined to quantify the coupling effect and can be expressed as

$$CCR = \begin{cases} \dfrac{|\beta^{'}(A) - \beta^{'}(B)|}{\beta^{'}(A)}, \text{ proximal bending} \\ \dfrac{\beta^{'}(A)}{\beta^{'}(B)}, \text{ distal bending} \end{cases} . \qquad (8)$$

The results indicate that ADC-V design achieves the largest bending angle during proximal and simultaneous bending, with deformation approaching that of ADC-A during distal bending, as shown in Fig. 5(d-i), (e-i), and (f-i). The *CCR* of ADC-V exhibits lower than that of ADC-A under proximal bending, while higher under distal bending. Although ADC-B design achieves the smallest $CCR$ during distal bending, as shown in Table 1, it exhibits the smallest bending angle. Consequently, the ADC-V is selected as the design paradigm.

Table 1. $CCR$ for 3 ADC designs

| $CCR$ | ADC-A | ADC-B | ADC-V |
|---|---|---|---|
| Proximal bending | 0.07% | 3.51% | 2.11% |
| Distal bending | 38.44% | 8.57% | 34.66% |

We discuss the geometric shape of the partition wall using three parameters: (1) the offset distance of the partition wall relative to the outer wall of the ventilation chamber, denoted as $d$; (2) the number of outer wall crease layers spanned by each crease of the partition wall, denoted as $g$; (3) the crease angle of the partition wall, denoted as $w$;, as illustrated in Fig. 5(c). For the AOB-H actuator, we tested its bending performance and motion decoupling capability by measuring the bending angles at plane A and B under three driving conditions: proximal bending, distal bending, and simultaneous bending, for different partition parameters, as shown in Fig. 5(d)-(f).

Plane A and plane B both exhibit comparable bending angle with different ADC designs and geometric parameters during proximal bending, which demonstrates the favorable decoupling performance of the ADC design for proximal bending, as shown in Fig. 5(d). During distal bending, the chamber coupling effect becomes pronounced, necessitating geometric parameter adjustment to suppress it, as shown in Fig. 5(e). During simultaneous bending, the bending angle of plane A and B show a consistent trend in response to changes in geometric parameters, as shown in Fig. 5(f). The parameter $d$ was varied in steps of 2. As $d$ increases, the bending angle decreases during proximal bending, while the bending angle first increases, then decreases, and subsequently increases again, reaching its maximum at $d = 14$ mm and $d = 6$ mm, respectively, as shown in Fig. 5 (d-ii), (e-ii) and (f-ii). The $CCR$ for proximal and distal bending are summarized in Table S3 in the SI 4. Notably, the $CCR$ for $d = 4$ mm, $d = 8$ mm and $d = 10$ mm remain below 25%. Parameter $g$ was varied in steps of 0.5. During proximal, distal and simultaneous bending, the AOB-H actuator achieved favorable bending angles at $g = 1$, as shown in Fig. 5 (d-iii), (e-iii) and (f-iii). Variations in $g$ do not induce significant changes in the $CCR$, indicating its limited influence on motion decoupling, as shown in Table S4. Parameter $w$ was varied in steps of $20°$. During proximal, distal and simultaneous bending, the AOB-H actuator achieved favorable bending angles at $w = 45°$ and $w = 65°$, as shown in Fig. 5 (d-iv), (e-iv) and (f-iv). The *CCR* for varied $w$ are summarized in Table S5. Based on a comprehensive evaluation of the actuator's bending angle range across three motion modes, the extent of coupling effects, and considering the manufacturing process, the design parameters for the ADC are finalized as listed in Table S6.

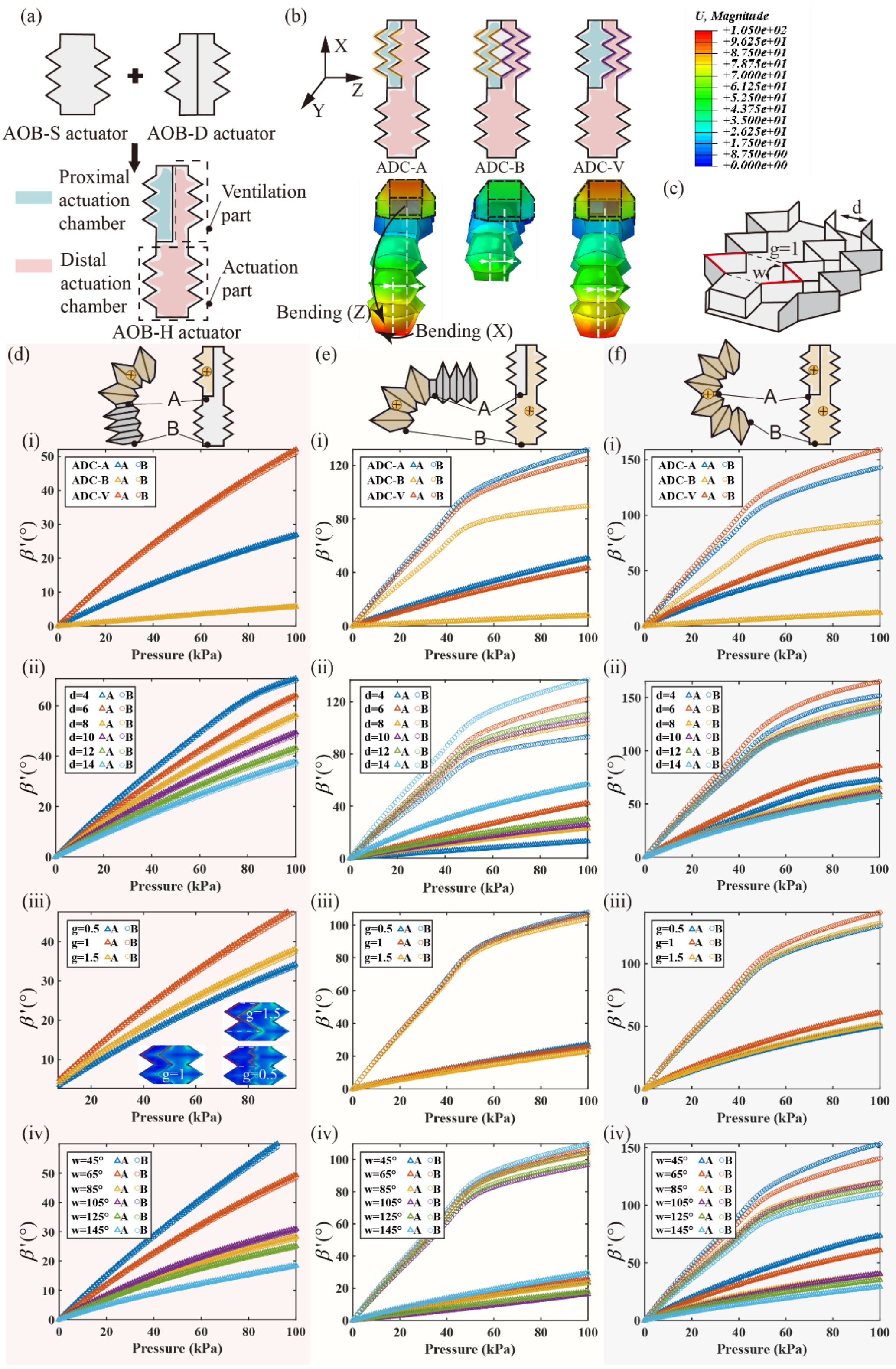

Fig. 5 Discussion on the ADC design. (a) Composition and chamber distribution of the AOB-H actuator. (b) Three types of ADC designs and corresponding FEA results, which indicate that under pressurization of the ventilation chamber, ADC-V produces the most uncoupled bending motion around x-axis. (c) Three design parameters of the partition. Bending performance of the actuator with different partition designs and parameters when only the proximal actuation chamber (d), distal actuation chamber (e) or the both chambers (f) are pressurized.

## IV. Design and fabrication of the OSOR hand

In this section, we present the detailed fabrication process and the design specifics of the OSOR hand.

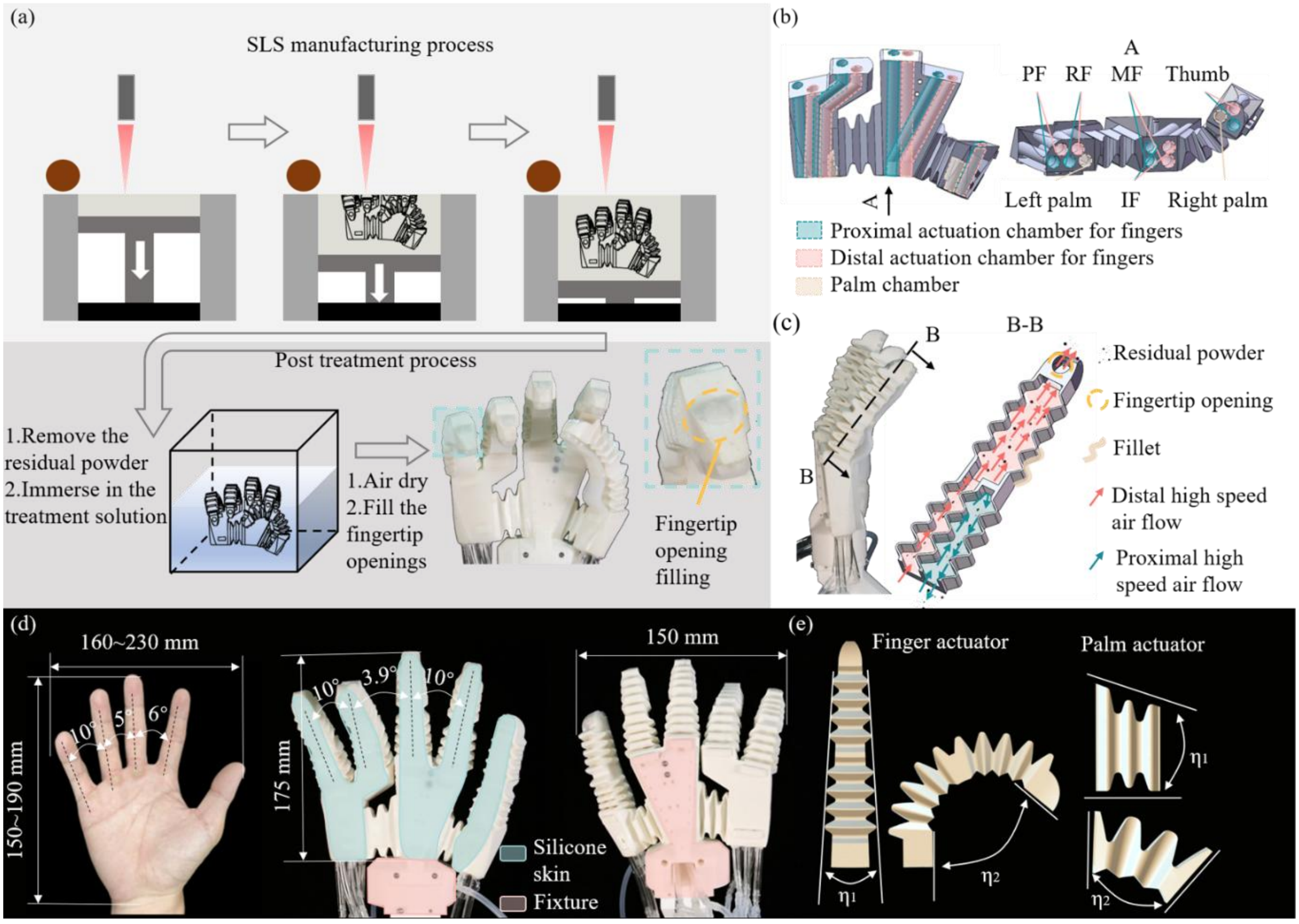

Fig. 6 Fabrication workflow and design details of the OSOR hand. (a) The OSOR hand is fabricated through SLS printing monolithically, followed by post treatment. (b) Optimized layout of the internal chambers within the palm is characterized by smooth curves and minimized sharp corners. (c) The process of powder removal from the soft robotic fingers after SLS printing. (d) The overview, fixture and silicone skin of the OSOR hand (e) Bio-inspired design in the finger and palm actuators.

To achieve single-material, monolithic fabrication of the OSOR hand, the AOB actuators are utilized to realize the desired motions in the fingers and the palm, while the supporting structures of the hand are designed to implement pneumatic channels within. To achieve intended motions and output forces from the origami actuators and the sufficient stiffness from the supporting structures, TPU is selected for the fabrication of the entire hand structure. When used to fabricate soft origami actuators [53], 95A TPU exhibits markedly higher payload capacity and superior energy-output efficiency, thus, is adopted for the fabrication.

To preserve design freedom and simplify post-processing, SLS 3D-printing was employed. As shown in Fig. 6(a), the OSOR hand was additively manufactured using a commercial SLS printer (EP-P3850, Shining 3D, Hangzhou). After printing, residual powders inside the prototype were removed from the openings at the fingertips and the wrist, as shown in Fig. 6(a) and (c), and then immersed in the surface treatment solution (FG-767 882, Fang Guan Industry) to form an airtight layer. The OSOR hand contains 12 air channels distributed as follows: 5 in the left palm for the pinky finger (PF), ring finger (RF) and left palm; 4 channels in the middle palm for the middle finger (MF) and index finger (IF); and 3 channels in the right palm for the thumb and right palm, as shown in Fig. 6(b). To ensure fluidity for the evacuation of powders and treatment liquids, the internal channels within the palm were designed with a cylindrical profile, and the creases of the fingers were smoothed using fillet transition to minimize sharp internal edges, as shown in Fig. 6(b) and (c). The printing parameters and immersion duration are detailed in the preceding work [53]. Finally, the OSOR hand was dried and the fingertip openings were sealed.

The OSOR hand was designed to approximate the dimensions of an average human hand as shown in Fig. 6 (d) and (e). For adult males, the hand span approximately ranges from 160 mm to 230 mm, and the hand length from 150 mm to 190 mm. In a natural relaxed posture, the interdigital angles are about 10° between the PF and RF, 5° between the RF and MF, and 6° between the MF and IF [35]. Considering these reference values, the final dimensions of the OSOR hand were determined to be 175 mm in span and 150 mm for length, with interdigital angles constrained within 10°. A fixture using FDM and customized silicone skin for the left, middle and right palm were fabricated and assembled to the OSOR hand via integral slots to support subsequent experiments and improve grasping performance, respectively, as shown in Fig. 6(d). Based on Table S3 and Table S7, structure bending angles $\eta_1$ and $\eta_2$ were incorporated in the finger and palm actuators to mimic the relaxed posture of human hand. $\eta_1$ represents the angle between tangent lines at the lateral mountain creases, and $\eta_2$ represents the bending angle of the actuator tip relative to its base, as illustrated in Fig. 6(e). The specific values of $\eta_1$ and $\eta_2$ for both the finger and palm actuators are listed in Table S8.

## V. OSOR Hand Characterization

In this chapter, we will verify the accuracy of the proposed model and test the output performance of the OSOR hand, including dexterity and grasping performance.

### A. Modelling validation and deformation tests

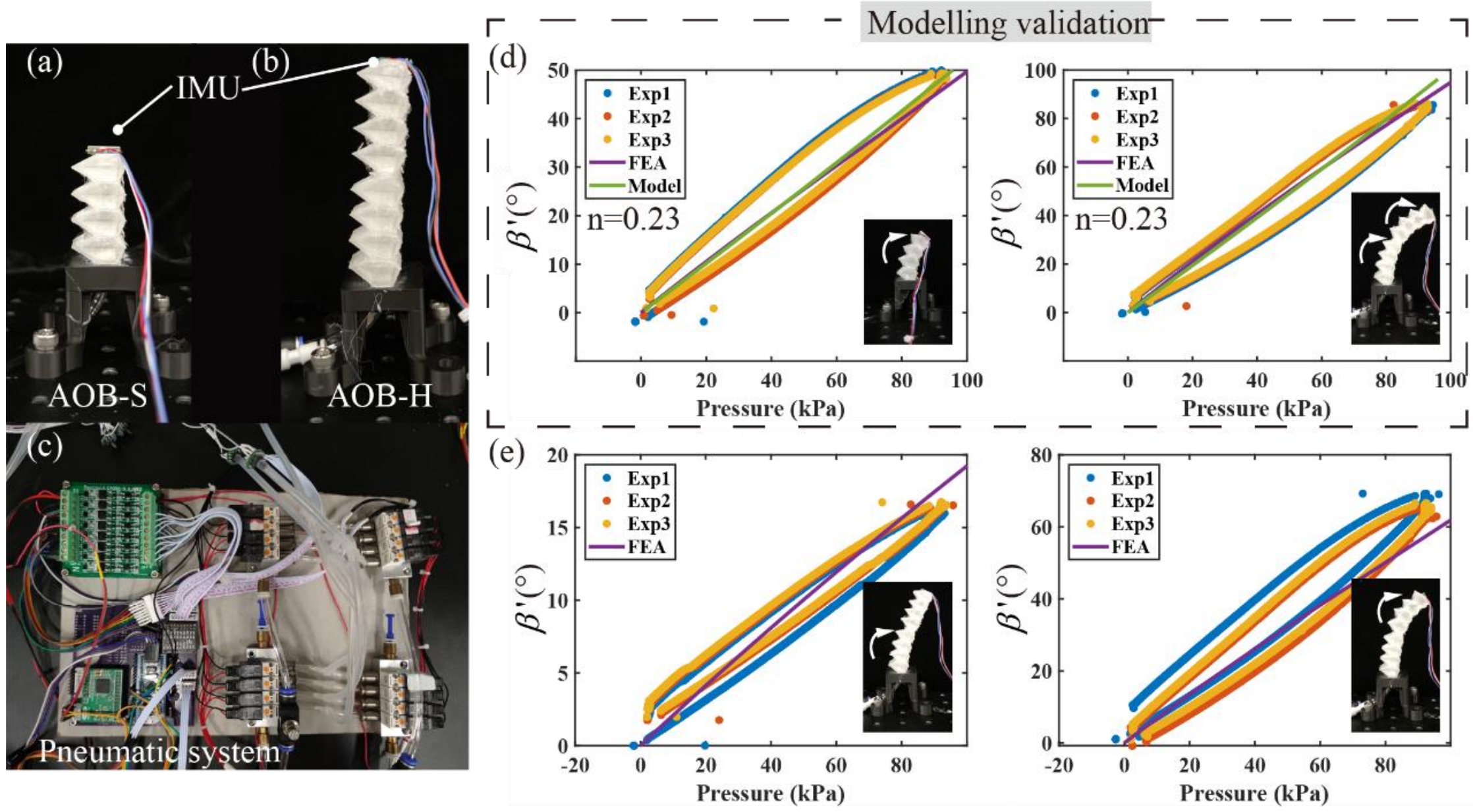

Fig. 7 Modelling validation. (a) The AOB-S actuator and (b) AOB-H actuator is attached with an IMU at the end. (c) The pneumatic system. (d) Modelling validation for the AOB actuator. (e) The motion performance of the AOB-H actuator during proximal bending and distal bending.

To validate the proposed model and evaluate deformations, we employed Fused Deposition Modeling (FDM) 3D-printing to rapidly fabricate the AOB-S and AOB-H actuators with TPU 95A with the parameters listed in Table S3 and Table S7, as shown in Fig. 7(a)-(b). The experiments on the AOB actuators were conducted using a pneumatic actuation system and an Inertial Measurement Units (IMUs) fixed on the distal ends of actuators. The pneumatic system is designed based on the Pump-valve method through a microcontroller (STM32f103c8t6, ST Inc.) to regulate two solenoid valves for each pneumatic channel. By gradually increasing the air pressure and synchronously recording the pressure and bending angle data, the relationship was obtained and plotted in Fig. 7(d) and (e).

As shown in Fig. 7(d), according to Eqns. (10) and (13), when $n$ is set to 0.23, the proposed model and FEA results effectively capture the motion trend of both actuators, with maximum bending angles reaching approximately 50° and 85°, respectively. Due to the hysteresis of soft material, the actuators exhibit a hysteresis loop between its loading and unloading process. The discrepancy between model theoretical results and empirical data arises from the inability to achieve a perfectly quasi-static process in experiments. In the model analysis, the lower stiffness of the actuator is primarily attributed to the omission of stretching and torsional deformation on some surfaces of the actuator in the model. For the proximal and distal bending motions of the AOB-H actuator shown in Fig. 7(e). The discrepancy between the FEA and experimental data primarily stems from the boundary condition settings. In the FEA model, when pressure is applied to the distal actuation chamber, its end face is treated as a rigid plane capable only of displacement without deformation. In experiments, however, this end face undergoes deformation, an effect not captured by the FEA. Consequently, during the proximal bending process, the stiffness predicted by the FEA is slightly lower than the experimental results; conversely, under the distal bending, the stiffness predicted by the FEA is higher than the experimental values. Nevertheless, the FEA results exhibited trends consistent with the experimental data. The experimentally measured maximum bending angles were approximately 17° and 71°, respectively.

B. Kinematic and static characteristics

The output performance tests of OSOR hand are shown in Fig. 8. The experiments employed the Arduino Mega 2560 as the core control board, with MATLAB programming interface used for command interaction to control the movements. The system is equipped with two air sources. The DAERTUO DET550-5L served as the positive pressure generator, while the DAERTUO oil-free vacuum pump functioned as the negative pressure generator. The two devices worked in coordination, and the stability of the air pressure was achieved. Through the using of IMU and force sensor, the bending angles and output forces of the actuators are obtained.

As shown in Fig. 8(a) and (b), an IMU was stuck to the corresponding palm actuators and finger actuators, respectively. When the pressure range was set to -50 to 140 kPa, with the horizontal position defined as 0°, the bending angle of the left palm changed from 40° to 68°, resulting in a bending angle change of 28°; the bending angle of the right palm changed from -20° to 20°, resulting in a bending angle change of 40°. Due to the manufacture effect, the IF, MF, RF and LF had slightly different bending angles in the unloaded state ($\eta_2$), which was around 88° to 93°. When pressure was applied simultaneously to the proximal and distal actuation chamber, the fingers on the same palm exhibited very similar pressure-angle variation trends. For the LF and RF, the bending angle changed from 46° to 198° with variations in pressure, corresponding to a bending range of 152°. For the MF and IF, the bending angle changed from 20° to 187° with variations in pressure, corresponding to a bending range of 167°; For the thumb, the maximum bending angle reached 188°, while under a negative pressure of -50 kPa, the bending angle reached -15°. Thus, the effective range of motion of the thumb is 203°.

The output forces at the fingertips of each finger and the palm were measured. The experimental setup is shown in Fig. 8 (c), where both the soft hand and the force sensor were placed horizontally to accurately measure the endpoint interaction force. When both actuation chambers were simultaneously given 140 kPa positive pressure and -50 kPa negative pressure, the performance of the fingers was as follows: the output force range for the right palm and left palm were 15 N and 16 N, respectively; the PF and RF achieved output force range of 5.1 N; MF, IF, and thumb had maximum output forces ranging between 6.3 N. In summary, the maximum output force for the palm reached 16 N, the maximum output force for a single finger reached 6.3 N, demonstrating promising force output performance, as shown in Fig. 8 (d).

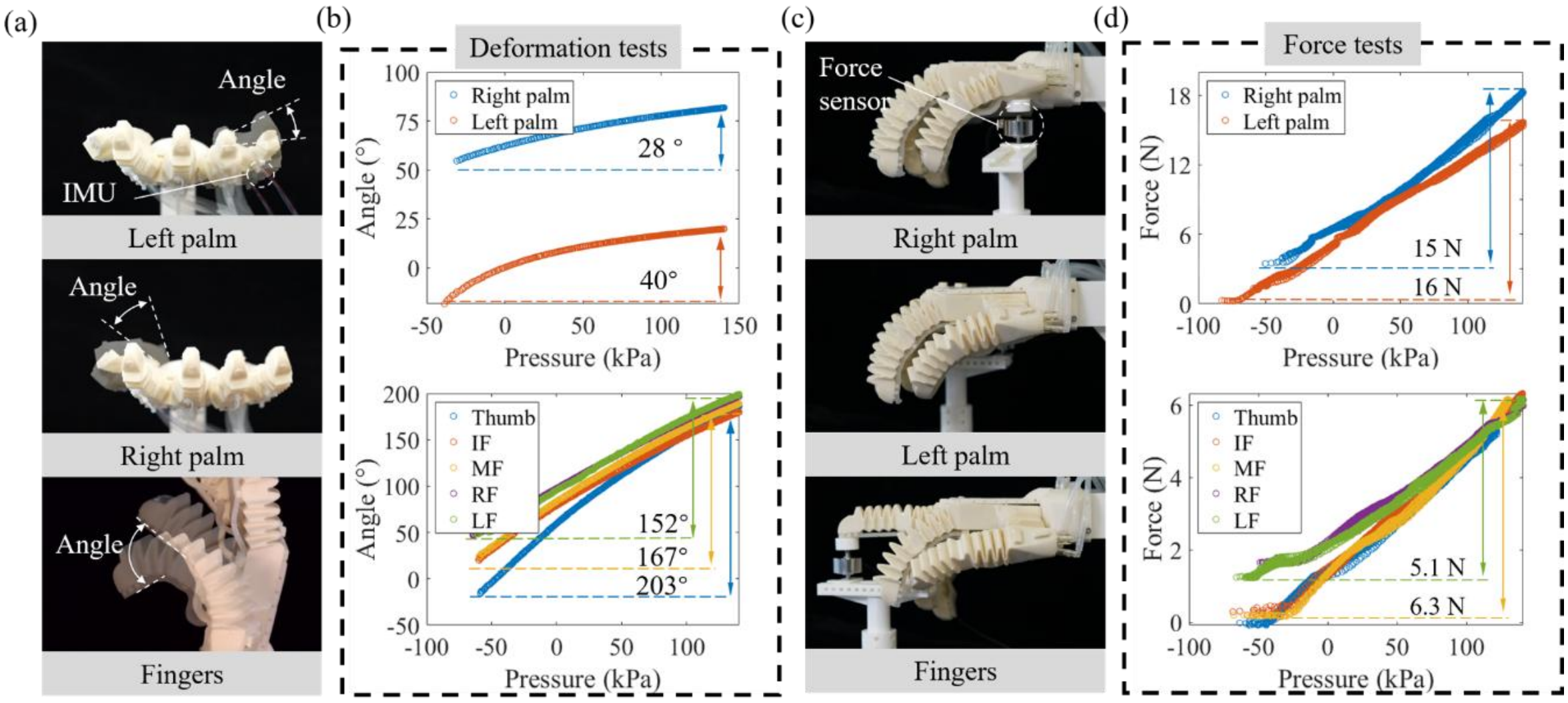

Fig. 8 Kinematic and static characteristic of the OSOR hand. (a) The experimental setup for the bending angle tests. (b) The bending angles of palm actuators and finger actuators. (c) The experimental setup for quantifying the force tests. (d) The output force of palm actuators and finger actuators.

### C. Flexibility

To test the flexibility of the OSOR hand, the 12-DOF motions performance of the OSOR hand were tested as shown in Fig. 9. When 100 kPa pressure was applied to the assigned air chamber, the corresponding finger joint exhibited bending displacements exceeding 90°, while the palm joint achieved displacements over 25°, as shown in the Fig. 9(a). We selected the joints of one finger from each of the left, middle, right palms, and the palm joints for illustration. The other joint motions are demonstrated in the Supplementary Movie S1.

Subsequently, the OSOR hand displayed a serious of thumb-to-fingertip contact maneuvers as shown in Fig. 9(b), providing evidence of its capability to execute tasks requiring dexterity. Additionally, by coordinating movements of the fingers and palm, the hand demonstrated Chinese gestures representing numbers from 0 to 9 as shown in Fig. 9(c). The process of the fingertip touching and gestures display can be referred to in the Supplementary Movie S2 and S3, respectively.

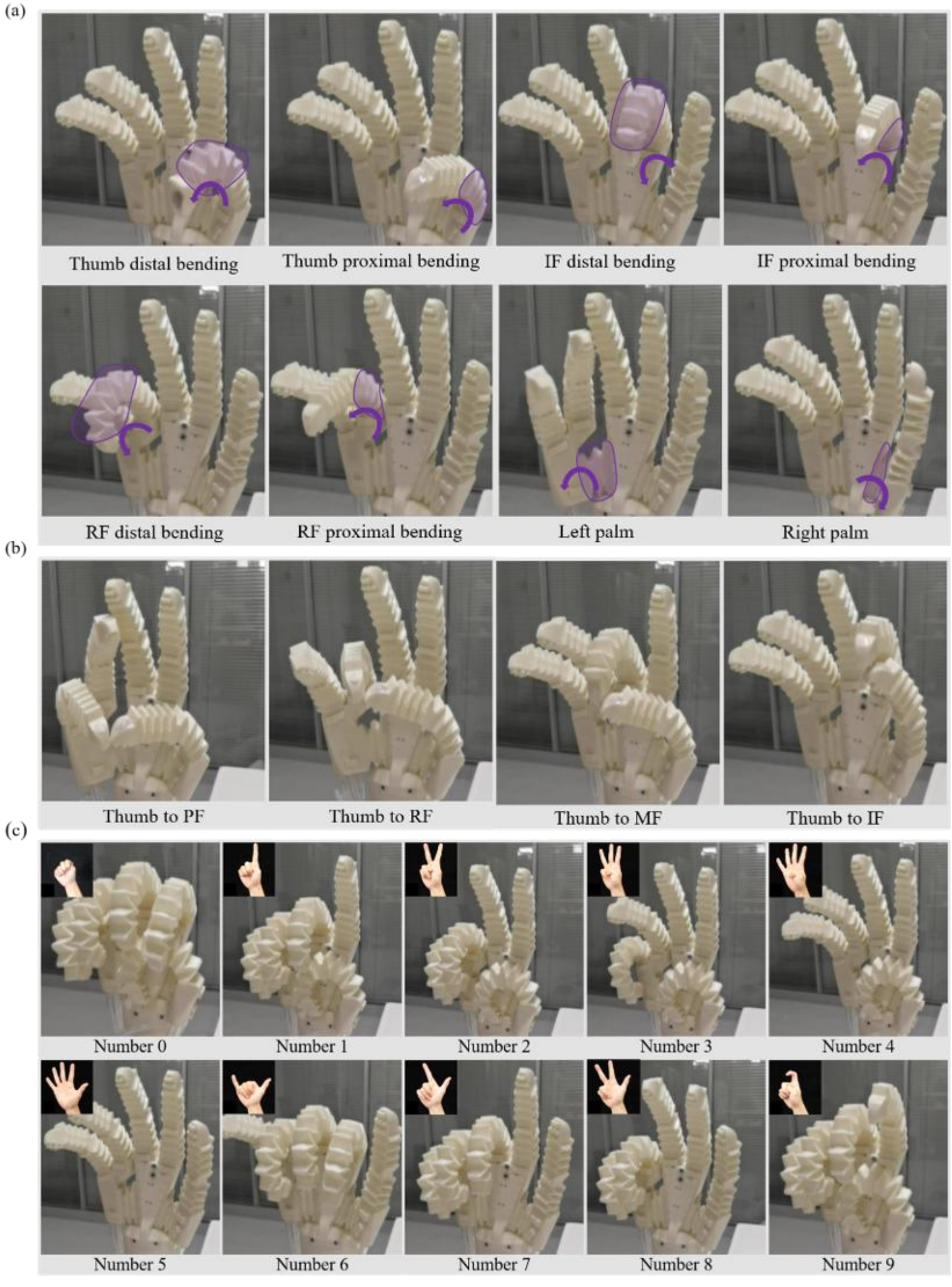

Fig. 9 Flexibility demonstrations for the OSOR hand. (a) Independent motions for the finger actuators and palm actuators. (b) The thumb tip touches the other four finger tips. (c) Representation of Chinese number gestures from 0 to 9.

D. Grasping capability

The grasping capability of each finger was firstly tested in the experiments using the sphere with a diameter of 60 mm, as shown in Fig. 10(a). The experimental results indicate each finger of the OSOR hand is capable of stably grasping the sphere exceeding 30 seconds. We then conducted tests using all the power grasps within the taxonomy of manufacturing grasps [94] as shown in Fig. 10(b). During the tests of Large Diameter and Medium Wrap gestures, the capability of the hand to grasp objects using only two fingers was tested. The hand successfully completed the tasks. Complementary to these foundational tests, the hand demonstrated two-finger coordination capability in Fig. 10(c): it reliably pinched delicate objects including paper tissues and a slender stick, while reliably grasping spherical items without slippage using multiple thumb-finger combinations such as thumb to IF, thumb to MF and thumb to PF, and performed controlled rolling manipulation using the fingertips. Two-finger coordinated grasping process was demonstrated in the Supplementary Movie S4. It further achieved unconventional grasping mode through palm-only motion coordination in Fig. 10(d), bypassing finger articulation entirely, as demonstrated in the Supplementary Movie S5. Practical functionality was verified in Fig. 10(e) via reliable grasping of complex daily tools such as a screwdriver and a glue gun, confirming consistent adaptation across object geometries (planar/curved/textured) and variable weights, which can be referred to in the Supplementary Movie S6. The OSOR hand also demonstrated the ability to take up and put down the household things including soft or tough things through the coordination of fingers and palms, such as a toy and a conch, which can be referred to in the Supplementary Movie S7.

In these tests, the OSOR hand demonstrated dependable grasping capabilities: it achieved two-finger grasps of cylindrical objects up to 100 mm in diameter (large water bottles) weighing over 1 kg; executed delicate pinch operations on sub-0.1 mm-thin fragile paper piece; and accomplished unconventional palm-only grasping on 85 mm diameter cylindrical vessels (beakers).

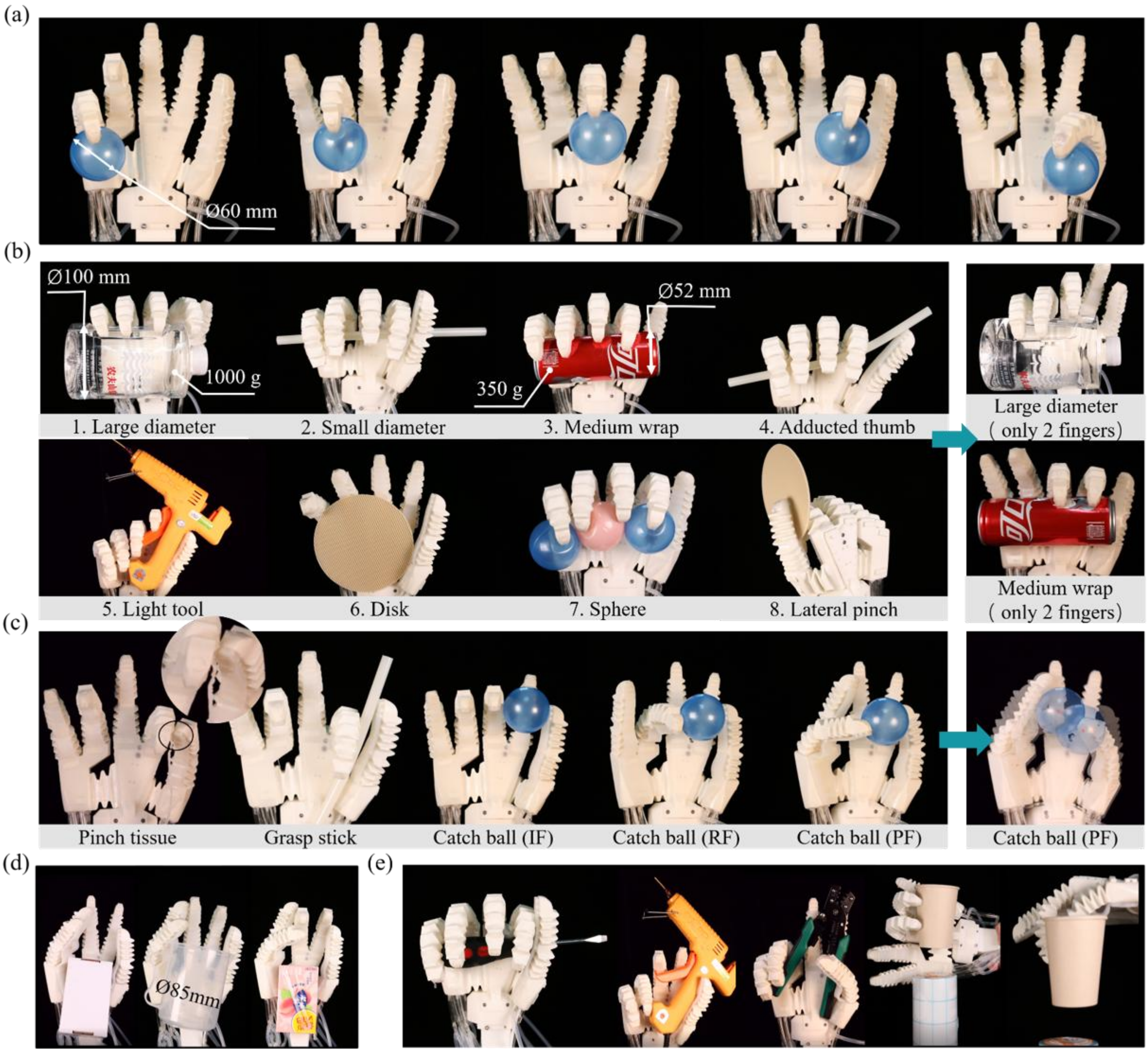

Fig. 10 Evaluation of grasping capabilities of the OSOR hand. (a) Single-finger grasping capability tests. (b) Gesture tests for power grasps within the taxonomy of manufacturing grasp [94]. (c) Two-finger coordinated grasping tests. (d) Palm-only grasping tests. (e) Grasping demonstration of common tools in daily life.

D. Comparison with existing soft robotic hand

A comparative analysis between the OSOR hand and state-of-the-art studies highlights its distinctive advantages. As summarized in Table 2, this work introduces a novel bending actuation principle for soft robotic hands through asymmetric dual-chamber origami design. This design enables monolithic fabrication of the entire hand, including five fingers and the palm. When compared with representative studies from the past six years, the proposed demonstrates top-tier performance in key metrics, particularly in terms of finger output forces and range of motions, positioning it at the forefront of current research achievements in this domain.

Table 2 Comparison with existing work

| Studies (Year) | Bending principle | Material | Fabrication | Assembly or not | Output force | Finger bending angle |
|---|---|---|---|---|---|---|
| Shorthose et al. (2022) [35] | IACA | Vero and Agilus30 | Multi-material polyjet | Y | 1.5 N | 186.3° |
| Sankar et al. (2025) [36] | IACA | PLA and silicone | 3D printing and molding | Y | 1.8 N | 208° |
| Zhou et al. (2018) [44] | IACA | PLA, silicone and fiber | 3D printing and molding | Y | 8.5 N | 86° |
| Zhou et al. (2019) [44] | IACA | Soft and rigid materials | Molding | Y | - | About 160° |
| Hao et al. (2022) [43] | SLA | ABS, fiber and silicone | 3D printing and molding | Y | 1.29 N | About 164° |
| Wang et al. (2021) [39] | SLA | Silicone and silkscreen fabric | Molding | Y | 1.23 N | 180° |
| Ren et al. (2023) [40] | SLA | Silicone and TPU | 3D printing and molding | Y | - | About 160° |
| Subramaniam et al. (2020) [41] | SLA | Silicone, ABS and fabric | 3D printing and molding | Y | 6~7 N | 110°~120° |
| Zhen et al. (2023) [46] | SLA | Silicone, nylon and fiber glass | 3D printing and molding | Y | 5 N | About 270° |
| **This work** | **AOB + ADC** | **TPU only** | **SLS only** | **N** | **Finger 6.3 N; Palm 16 N** | **Finger 203°; Palm 40°** |

# VI. Conclusion

This paper proposed the AOB pattern and the ADC design for achieving bending motions and multi-functionality for soft robotic hands based on the Yoshimura origami prismatic structure, respectively.

By defining the asymmetric ratio $R$ to capture the relative distance between the middle and end plane, the geometric asymmetry of the AOB unit is defined, which generates the bending motions of the AOB actuators. By analyzing the mechanism of coupled deformation caused by the asymmetric foldable partition, the optimized ADC-V was developed for the AOB actuators. The combination of AOB pattern and ADC design endows the AOB-D chamber with a ventilation function in addition to its primary actuation capability. Through connecting AOB-S chamber and the AOB-D chamber, the AOB-H chamber was obtained, thereby forming a 2-DOF finger actuator and the AOB-S chamber was utilized as the palm actuators. The OSOR hand with 5 fingers and the palm driven by AOB actuators was developed with 12 DOFs. The design enabled monolithic fabrication of the hand from TPU via SLS process, eliminating the need for assembly.

To guide the design, analytical models were established for the AOB actuator. Ratio $R$ endows the AOB actuator with programmable geometric and motion characteristics. The parametric analysis of the AOB pattern and ADC design was carried out, together with extensive FEAs, to investigate the influence of geometric parameters on the motion performance and to determine the parameters for the OSOR hand.

Validation of the analytical model showed that the bending angles of AOB actuators, including AOB-S and AOB-H actuators, were successfully predicted. Experimental validation confirmed the hand's exceptional performance. The palm actuators exhibited maximum bending angle ranges of 28° and 40°, respectively, while the finger actuators showed maximum ranges between 152° and 203°. The palm actuators achieved a maximum output force of 16 N, while the finger actuators achieved a maximum fingertip force of 6.3 N. Dexterity tests demonstrated the hand's ability to perform preset 12-DOF motions, successfully executing 4 distinct fingertip-to-fingertip opposition gestures and representing Chinese numerical gestures from 0 to 9. In grasping tests, the hand was proved to be capable of power grasps within the taxonomy of manufacturing grasps and to securely manipulate a wide range of daily objects using comprehensive finger-palm coordinated motion strategies.

In summary, this work provides an asymmetric dual-chamber origami design method for obtaining hand’s bending motions, and a framework for designing and fabricating high-performance, monolithic soft robotic hands. The OSOR hand demonstrates that complex functionality can be encoded into the structure itself, streamlining the manufacturing process. Future work will focus on enhancing the dexterity, integrating sensory feedback systems, and developing advanced control strategies to tackle complex manipulation tasks.